# ANGLE CONSTRAINED PATH TO CLUSTER MULTIPLE MANIFOLDS

Amir Babaeian

University of California, San Diego

Abstract. In this paper, we propose a method to cluster multiple intersected manifolds. The algorithm chooses several landmark nodes randomly and then checks whether there is an angle-constrained path between each landmark node and every other node in the neighborhood graph. When the points lie on different manifolds with intersection they should not be connected using a smooth path, thus the angle constraint is used to prevent connecting points from one cluster to another one. The resulting algorithm is implemented as a simple variation of Dijkstra's algorithm used in Isomap. However, Isomap was specifically designed for dimensionality reduction in the single-manifold setting, and in particular, cannot handle intersections. Our method is simpler than the previous proposals in the literature and performs comparably to the best methods, both on simulated and some real datasets.

## 1 Introduction

Since last decade, the amount of collected data is increasing exponentially. Data mining techniques represent each data sample with an extracted very high-dimensional fea-ture vector. However, the intrinsic dimension of these features is much smaller that the dimension of ambient space. Precisely, the features are sampled from the surface of a low-dimensional manifold embedded in the ambient space [7, 19]. The most well-known manifold learning algorithms such as Locally Linear Embedding (LLE) [16], Laplacian Eigenmap (LE) [2], and Isomap [18]. All these methods assume that the data is coming from a single manifold. However, this assumption is not correct in many application such as video segmentation [5], where the data is sampled from multiple intersected manifolds. However, clustering the data points into several intersected man-ifolds is itself a challenging problem.

In this work, we consider the problem of clustering data points sampled from the vicinity of multiple intersected manifolds embedded in ambient space. Here, we focus on a nonparametric approach based on the assumption that the manifolds are smooth. See Fig. 1.

Related works in the area are based on different assumptions. Most methods are designed for the cases where the surfaces do not intersect [15, 14, 6], while others work only when the surfaces that intersect have different intrinsic dimension or density [8, 12]. The method of [1] is only able to separate intersecting curves. Methods that pur-posefully aim at resolving intersections are fewer. [17] implement some variant of K-means where the centers are surfaces. [11] propose to minimize a (combinatorial) en-ergy that includes local orientation information, using a tabu search. The state-of-the-art methods work based on local principal component analysis (PCA). An early proposal was the elaborate multi-scale spectral method of [13], while the clustering routine of [9], developed in the context of semi-supervised learning, inspired by the works of [20] and [10]. Here, we propose a markedly different approach based on connecting points to landmarks via angle-constrained paths. It can be seen as a constrained variant of Isomap [18]. Isomap was specifically designed for dimensionality reduction in the single-manifold setting, and in particular, cannot handle intersections. The angle con-straint on paths is there to prevent connecting points from one cluster to points from a different, intersecting cluster. The resulting algorithm is implemented as a simple varia-tion of Dijkstra's algorithm. The experimental results, conducted on both synthetic and real real data sets, confirm that our method performs comparably to the state-of-the-art methods.

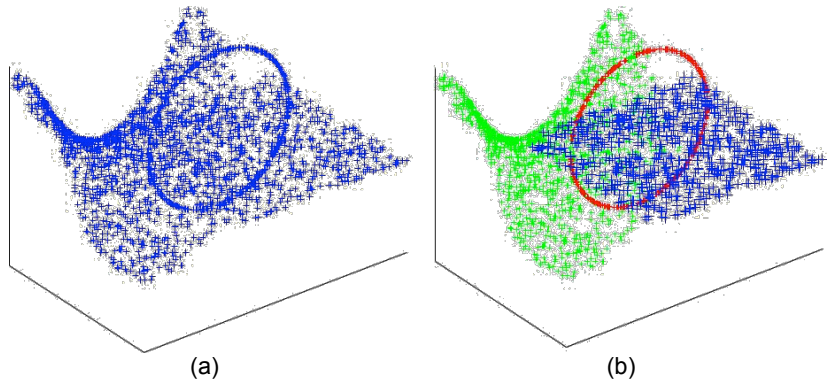

Fig. 1. Simulated data illustrating the problem of multi-manifold clustering; (a) A sample 3D data containing three intersected manifolds; (b) three clustered manifolds as the output of the proposed method.

The rest of the paper is organized as follows. In Section 2, we describe in detail the problem of multi-manifold clustering followed by discussing the methods of [17] and [20], as well as the subspace clustering method of [4], which will serve as bench-marks in our experiments. In Section 3 we motivate and describe our proposed method. In Section 4, we present some numerical experiments on synthetic and real data sets. Finally, in Section 5, we draw our conclusion.

## 2  Multi-Manifold Clustering

We observe points $x_1, \ldots, x_N$ in a Euclidean space $R^D$, with $D \geq 2$, that are assumed to be sampled from the vicinity of embedded submanifolds $S_1, \ldots, S_K \to R^D$. The surfaces are assumed to be compact and smooth — defined here as having bounded maximum pointwise curvature and a boundary (if any) satisfying the same property — and may be of possibly different intrinsic dimensions. The points are independent and identically drawn from a mixture distribution supported in the neighborhood of the

surfaces. A simple model assumes

$$x_i = s_i + z_i, \quad s_i \leftarrow \sum_{k=1}^{K} \pi_k \mu_{S_k}, \quad \|z_i\| \leq \epsilon, \quad (1)$$

where $\mu_S$ denotes the uniform distribution over S. ($s \leftarrow \mu$ means that the random point s has distribution μ.) The mixture weights satisfy $\pi_i \geq 0$ and $\sum_k \pi_k = 1$, and $z_i$ is a uniform noise bounded by $\epsilon$. Let $I_k$ index the points $s_i$ that were sampled from surface $S_k$, meaning $I_k = \{i : s_i \leftarrow \mu_{S_k}\}$. Note that $I_1, \ldots, I_K$ forms a partition of the data. (This implicitly assumes there are no outliers, which we take to be the case in this pa-per.) The goal of multi-manifold clustering is to recover this partition up to permutation of $\{1, \ldots, K\}$. We note that the term 'submanifold' is unduly restrictive for us since we allow the underlying surfaces to self-intersect. Our method works essentially as well on such surfaces as on sub-manifolds. See 2.

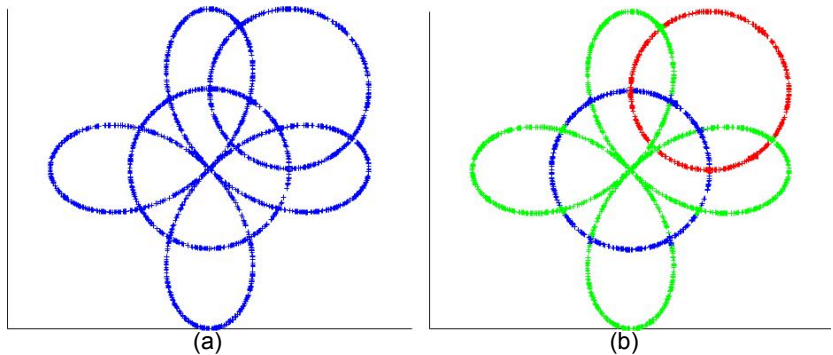

(a)    (b)

Fig. 2. Simulated data illustrating our method working well in a case where the underlying sur-faces self-intersect. Left: data. Right: output from our method.

We also note that, when the underlying surfaces are allowed to intersect, some sort of smoothness assumption seems necessary in a nonparametric setting like ours, for otherwise the clustering objective is ambiguous — think of a T-intersection.

Authors in [17] suggest an algorithm that mimics K-means algorithm by replacing centroid points with centroid sub-manifolds. The method starts like Isomap by build-ing a neighborhood graph and computes the shortest path distances within the graph. In [20], dissimilarity is used to factor the Euclidean distance and the discrepancy be-tween the local orientation of the data. The surfaces are assumed to be of same dimen-sion d known to the user. We chose the subspace clustering method of [4] among a few other methods that perform well in this context.In [4], people proposed a spectral method for subspace clustering based on the affinity of underlying manifolds.

# 3 Approach

Our algorithm is quite distinct from all the other methods for multi-manifold clustering we are aware of, although it starts by building a q-nearest neighbor graph like many others. The idea is very simple and amounts to clustering together points that are connected by an angle-constrained path in the neighborhood graph. For an angle $\checkmark \in [0, \pi]$, we say that a path $(x_i^1, \ldots, x_i^m)$ is $\checkmark$-constrained if $\angle(x_i^{t-1} x_i^t, x_i^t x_i^{t+1}) \leq \checkmark$ for all $t = 2, \ldots, m-1$.

The rationale is the following. Take two surfaces $S_1$ and $S_2$ intersecting at a strictly positive angle. Then for 'most' pairs of data points $x_{i1} \in S_1$ and $x_{i2} \in S_2$, a path in the graph going from $x_{i1}$ to $x_{i2}$ has at least one large angle between two successive edges, on the order of the incidence angle between the surfaces; while for 'most' pairs of data points $x_{i1}, x_{i2} \in S_1$, there is a path with all angles between successive edges relatively small.

To speedup the implementation, we select M landmarks (with M slightly larger than K) at random among the data points and only identify what data points are con-nected to what landmark via a $\checkmark$-constrained path in the graph. M and $\checkmark$ are parameters of the algorithm.

Let $\xi_{\ell i} = 1$ if point i and landmark $\ell$ are connected that way, and $\xi_{\ell i} = 0$ if not. We use $\xi_i := (\xi_{\ell i} : \ell = 1, \ldots, M)$ as feature vectors that we group together and cluster using hierarchical clustering with complete linkage.

## 3.1 Computational Complexity

Building a symmetric q-nearest neighbor graph using cover trees [3] takes order $O(qN \log N)$, where the implicit constant depends exponentially on the intrinsic dimensions of the surfaces and linearly on the ambient dimension D. The angle-constrained pathfinder routine is a simple variant of Dijkstra's algorithm, whose implementation by Fibonacci heaps runs in $O(qN \log N)$. Hence, calling this routine once for each landmark costs $O(qM N \log N)$. Grouping the feature vectors $O(F M N)$ and then clustering them by complete linkage costs $O(F^2 \log F)$, where F is the (data-dependent) number of distinct feature vectors $\xi_i$, often of the order of K in our experiments.

---

**Algorithm 1** Path-Based Clustering (PBC)

---

Input: data $(x_i)$; parameters q, K, M, $\checkmark$
Build q-nearest neighbor graph
Choose M landmarks are random
for i = 1 to n do
　　For each landmark $x^\ell$, identify which points $x_i$ it is connected to via a $\checkmark$-
　　constrained path in the graph, and set $\xi_{\ell i} = 1$ if so, and $\xi_{\ell i} = 0$ otherwise.
end for
Group and then apply hierarchical clustering to the feature vectors $\xi_1, \ldots, \xi_n$ to
find K clus-ters, where $\xi_i := (\xi_{\ell i} : \ell = 1, \ldots, M)$.

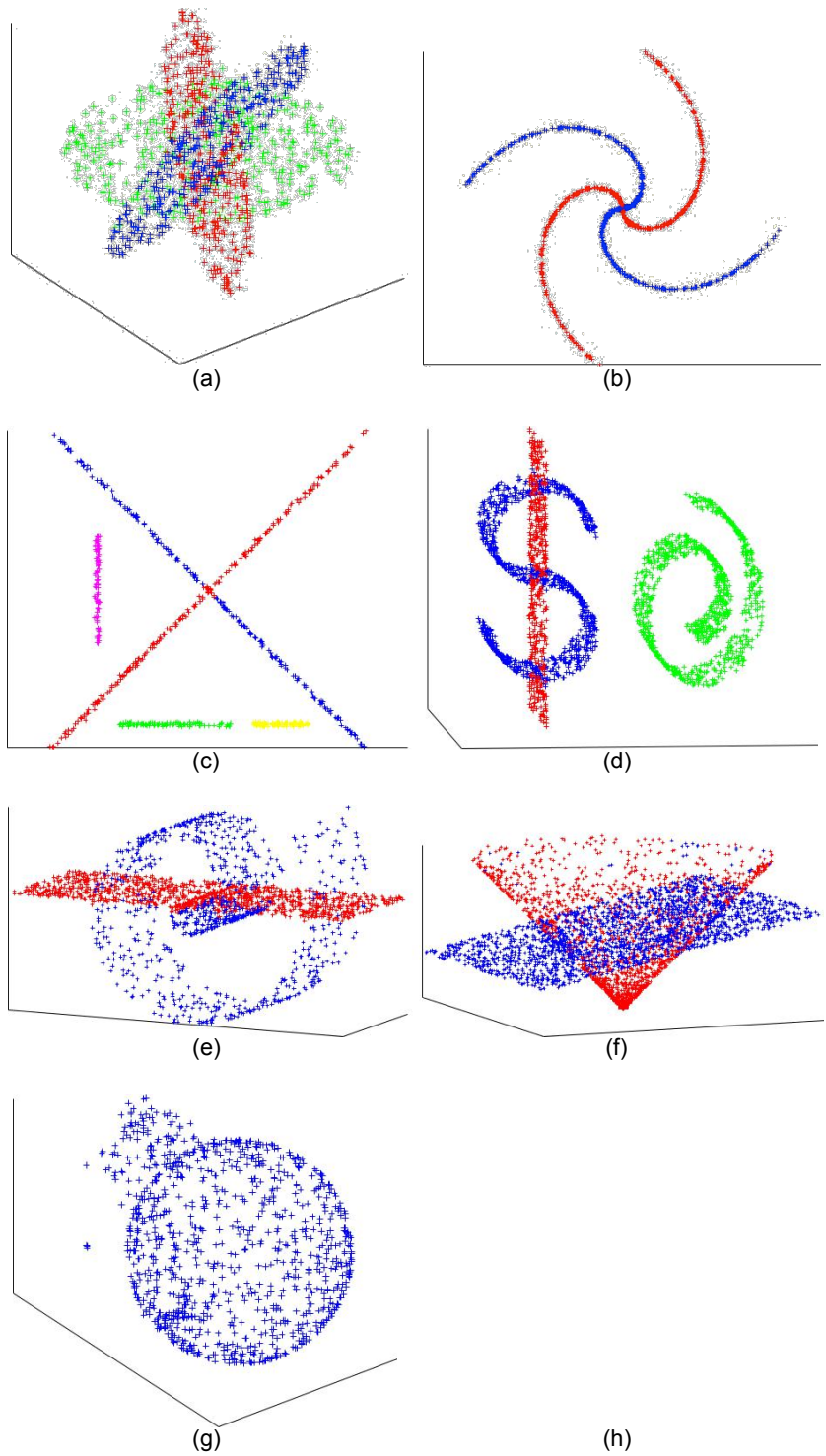

Fig. 3. Result of our method on 8 synthetic datasets.

## 4 Experiments

In this section, we compare the performance of our methods with those described in 2 on various synthetic and real data sets.

### 4.1 Synthetic Data

The synthetic datasets we generated are similar to those appearing in the literature. They are Three Planes (TP), Two Spirals (TSI), Five Segments (FS), Dollar-Sign, Plane and Roll (DSPR), Roll and Plane (RP), Cone and Plane (CP), Two Spheres (TSH), Rose Curve and Circle (RCC). Figure 3 shows the result of our method on eight synthetic datasets, with good performance in all cases. The misclustering rates for our method, and the other three methods, are presented in Table 1, where we see that our method achieves a performance at least comparable to the best of the other three methods on each dataset. Spectral Curvature Clustering (SCC) works well on linear manifolds (as expected) while it fails when there is curvature. See Fig. 4(a). K-Manifolds fails in the more complicated examples. We found that this algorithm is very slow since it has to compute the shortest path between all the points, so that we could not apply it to some of the largest datasets. We mention that it assumes that clusters intersect, and otherwise does not work properly. Fig. 4(b) and Fig. 4(c) shows two cases where K-Manifolds fails. Our method and Spectral Multi-Manifolds Clustering (SMMC) perform compa-rably on most datasets, but SMMC fails in the Rose Curve and Circle example shown in 4(d). We note that K-Manifold, SCC and SMMC all require that all surfaces are of same dimension, which is a parameter of these methods, why our method does not need knowledge of the intrinsic dimensions of the surfaces and can operate even when these are different.

Table 1. Clustering accuracy on synthetic data.

| Data set | K-Manifolds | SCC | SMMC | PBC |
|---|---|---|---|---|
| TP | 95.1% | 98.8% | 99.3% | 93.0% |
| TSI | 95.2% | 54.8% | 99.6% | 98.2% |
| FS | 59.1% | 94.9% | 99.6% | 98.1% |
| DSPR | 46.1% | - | 99.4% | 97.5% |
| RP | 53.5% | - | 95.6% | 94.6% |
| CP | - | - | 99.3% | 94.9% |
| TS | - | - | 95.5% | 97.9% |
| RCC | 62.9% | - | 64.7% | 99.4% |

### 4.2 Real Data

We applied our method on the COIL-20 dataset which includes 1440 gray-scale images of the size 128 ↠ 128 pixels from 20 objects. Each object contains 72 images taken from different view points. We first projected the data set onto the top 10 principal components, then applied our path-based clustering algorithm. We tested our method on the three very similar objects 3, 6 and 19. The algorithm is 99% accurate (misclusters are only 2 images out of 216) bringing a significant improvement over the state-of-the-art result of 70% reported in [20]. Lastly, we evaluated our method on the whole dataset obtaining an 83.6% accuracy, improving on the 70.7% accuracy reported in [20]. Since in this case we have 20 different classes, we increased the number of landmarks to 100 to make sure we have sampled at least a few landmarks from each class. In 5, we show the results of our algorithm and those of SCC and SMMC on first 5 objects in a 3D plot where the points are represented by their first 3 principal components. As it can be seen other 3 methods fail to cluster multiple manifold with intersection whereas Path-Based Clustering seems much closer to the ground truth.

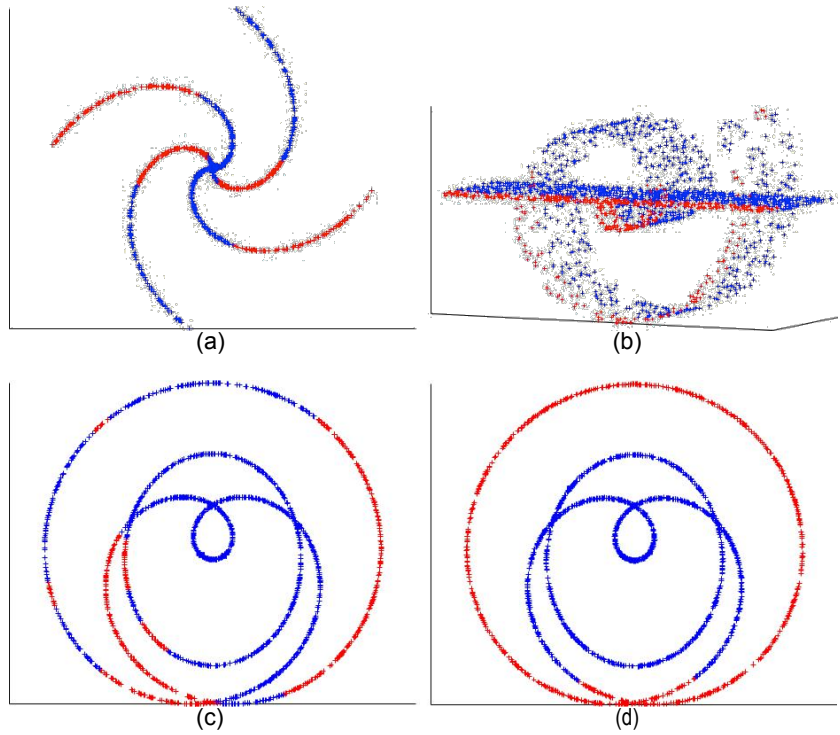

Fig. 4. (a) An example where SCC fails; (b-c) examples where K-Manifolds fails; (d) An example where SMMC fails.

### 4.3 Clustering of Human Motion Sequences

In this section, we test our algorithm on a sequence of video frames including differ-ent activities performed by a subject. [21, 22, 24, 23] studied unsupervised tracking of human bodies in the presence of temporal changes. We use a different model to track 4 mixed actions from subject 86, trial number 9 of the CMU MoCap dataset. The data consists in a temporal sequence of 62-dimensional representation of the human body via markers in R$_3$. One motion sequence of 4794 frames and corresponding result of path-based multi-manifold clustering are given in Figure 6. Four activities are labeled from 1 to 4. When the subject switches from one action to another one the labels are ambiguous.

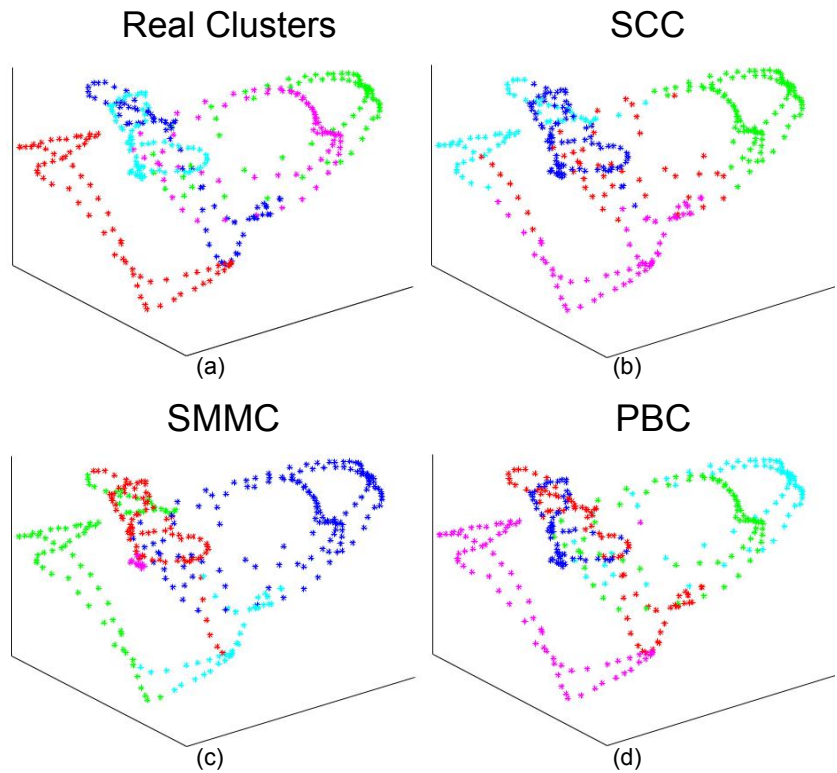

Fig. 5. Output of three different methods on the first 5 objects of COIL-20. Points are represented by their 3 top principal components.

## 5 Conclusion

In this paper, we proposed a novel method to cluster multiple intersected manifolds that perform well in practice and is consistent in the large-sample limit. The variant we introduced is the best on numerical experiments. For now, the variant we introduced here is the best on numerical experiments.

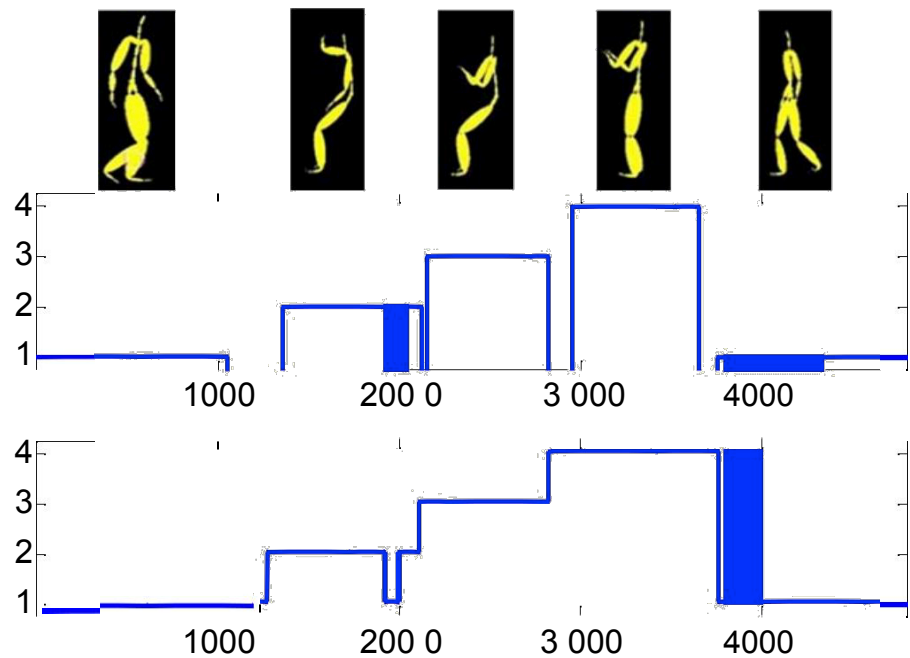

Fig. 6. Result of human activity segmentation using Path-Based Clustering. There are 4 activities: walking (1), looking (2), sitting (3) and standing (4). Top: a sample of the sequence. Middle: ground truth. Bottom: output of our algorithm.

## Acknowledgments

This work was partially supported by a grant from the National Science Foundation DMS- 09-15160.